# Minimizing Area and Energy of Deep Learning Hardware Design Using Collective Low Precision and Structured Compression

Shihui Yin, Gaurav Srivastava, Shreyas K. Venkataramanaiah, Chaitali Chakrabarti, Visar Berisha, and Jae-sun Seo

School of Electrical, Computer and Energy Engineering

Arizona State University, Tempe, USA

*Abstract*— Deep learning algorithms have shown tremendous success in many recognition tasks; however, these algorithms typically include a deep neural network (DNN) structure and a large number of parameters, which makes it challenging to implement them on power/area-constrained embedded platforms. To reduce the network size, several studies investigated compression by introducing element-wise or row-/column-/block-wise sparsity via pruning and regularization. In addition, many recent works have focused on reducing precision of activations and weights with some reducing down to a single bit. However, combining various sparsity structures with binarized or very-low-precision (2-3 bit) neural networks have not been comprehensively explored. In this work, we present design techniques for minimum-area/-energy DNN hardware with minimal degradation in accuracy. During training, both binarization/low-precision and structured sparsity are applied as constraints to find the smallest memory footprint for a given deep learning algorithm. The DNN model for CIFAR-10 dataset with weight memory reduction of 50X exhibits accuracy comparable to that of the floating-point counterpart. Area, performance and energy results of DNN hardware in 40nm CMOS are reported for the MNIST dataset. The optimized DNN that combines 8X structured compression and 3-bit weight precision showed 98.4% accuracy at 20nJ per classification.

## I. Introduction

Deep neural networks (DNNs) have seen great success in many cognitive applications such as image classification [1-2] and speech recognition [3]. However, the large number of operations and parameters in state-of-the-art DNN algorithms have posed significant challenges for energy-efficient DNN hardware designs. In particular, embedded hardware systems or edge computing devices are constrained by limited computing resources and memory, necessitating hardware implementations to use techniques to reduce the neural network size and lower energy consumption. A number of prior works investigated methods to (1) lower the precision of activations and weights [4-6] and (2) apply pruning and compression techniques [7] for DNNs while maintaining high classification accuracy.

Low-precision techniques rely on quantizing the DNN weights and activations with a small number of bits. The extreme case of DNN quantization is binarizing the weights and activations. BinaryConnect [4] showed that binarizing the weights does not adversely affect accuracy and, in some cases, can improve the accuracy, compared to non-binarized DNN. Binarized Neural Network (BNN) [5] extended the approach by binarizing both weights and activations and XNOR-Net [6] used binarized network for ImageNet classification.

Many prior works have also attempted to compress DNNs [7] through pruning of neurons and weights. However, generating a scattered sparsity may not necessarily result in acceleration on hardware [9, 10], and can also increase the storage overhead for encoding sparsity. Coarse Grain Sparsity (CGS) has been proposed in [8], where static sparsity is applied on randomly selected blocks of weights throughout training. Structured Sparsity Learning (SSL) [9] has demonstrated row-/column-/layer-wise structured sparsity based on group Lasso regularization, achieving 5.1X speedup on GPU compared to non-structured sparsity. Scalpel [10] customized DNN pruning to the underlying hardware by introducing structured sparsity that matches the data-parallel hardware organization.

While these prior works investigated low-precision and structured compression in isolation, there has been little work that systematically applied and optimized both techniques in a single framework. Deep compression [11] applied pruning and quantization on weights; however, the sparsity remained non-structured. Prior CGS work [8] employed block-wise structured sparsity, but only quantized the weights and activations after training was complete, resulting in limited precision reduction (5-6 bit). To simultaneously achieve very low precision (1-3 bit) and structured sparsity in DNNs, both of the techniques need to be applied throughout the training process and classification [4, 5, 8]. Only then, the overall reduction in memory and computation will be substantial while minimizing the index overhead that stores sparsity information, resulting in prominent acceleration with low-area/-energy hardware implementation.

In this work, we investigate jointly applying low-precision and structured sparsity constraints during DNN training, such that the DNN hardware for classification can be implemented with very low area and energy. Our main contributions are:

1) We applied weight and activation quantization for different bit-precision combinations, while applying Coarse-Grained Sparsity (CGS) [8] constraints. We studied the effect of quantization and structured-sparsity on weight memory usage, test accuracy, hardware energy and area requirements.

2) We implemented custom digital hardware for various combinations of low-precision and compression, and demonstrated low-area/-energy DNN hardware design.

3) The proposed methodology is empirically validated by implementing the inference phase of DNNs for MNIST and CIFAR-10 datasets. The CNN for CIFAR-10 achieves ~50X weight memory reduction with accuracy comparable to that of the floating-point CNN. The DNN for MNIST with 8-bit activations, 3-bit weights and 8X CGS compression showed 98.4% accuracy at 20nJ per classification, which is a >10X energy improvement compared to the baseline DNN.

## II. OVERVIEW OF LOW-PRECISION TECHNIQUES FOR DNNS

As the majority of computations in DNN are multiply-and-accumulate (MAC) operations, constraining the weights and activations to low precision during training can result in significant speedup with appropriate hardware design for classification. BinaryConnect [4] uses quantized value of the real-valued weights for forward and backward phases of back propagation. With the weights constrained to just 1-bit, the MAC operations can be replaced with simple additions and subtractions. BNN [5] quantized both weights and activations to +1 or -1, where MAC operations become bit-wise XNOR and accumulate operations. Authors in [4, 5] argue that the quantization noise acts like a regularizer and hence can give good test accuracy even with 1-bit quantization. BNN uses straight-through estimator [12] to approximate the gradient of quantized activations in backward propagation. Compared to BNN, XNOR-Net [6] showed large improvement in ImageNet classification accuracy with binary weights and activations.

## III. OVERVIEW OF STRUCTURED COMPRESSION TECHNIQUES FOR DNNS

### A. Structured Sparsity Learing

Structured Sparsity learning (SSL) [9] applies group Lasso regularization [13] to the weights belonging to a DNN structure (filters, channels, filter shapes, layer depth). This prunes the weights corresponding to the unimportant structures in the DNN model. SSL generates compact DNN structures, which can be efficiently implemented in hardware. Specifically, it applies sparsity constraints on 2D filters within a 3D filter resulting in significant reduction in the convolution-related computations. Also, filter-sparsity and shape-sparsity can be used to reduce the size of weight matrix. As shown in Fig. 1, the convolution operation in DNNs is usually converted to General Matrix to Matrix Multiplication (GEMM) by converting weight tensors and feature tensors to matrices [14]. Filter and shape sparsity can be used to remove rows and columns of this weight matrix.

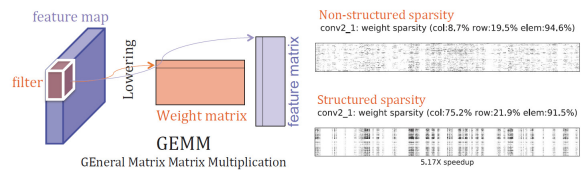

Fig. 1. Structured sparsity for GEMM speedup (adapted from [10]).

### B. Scalpel

Scalpel [10] applies DNN sparsity depending on the level of data-parallelism of the target hardware. Matrix multiplications on a sparse matrix need extra computations to decode the sparse format of the matrix. For low-parallelism hardware, SIMD-aware weight pruning maintains weights in aligned fixed-size groups to fully utilize the SIMD units. For high-parallelism hardware, node pruning is applied. For moderate-parallelism hardware, a combination of SIMD-aware weight and node pruning is performed.

### C. Coarse-Grain sparsity

Coarse-Grain sparsity (CGS) [8, 15] is a technique to generate structured sparsity by randomly dropping blocks of weights within the DNN weights matrix throughout training. The overall sparsity depends on the CGS block size and the CGS compression ratio (CGS ratio). Since sparsity is formed on a block-by-block basis, the index overhead is minimized, allowing the final trained weights to be efficiently mapped onto SRAM arrays.

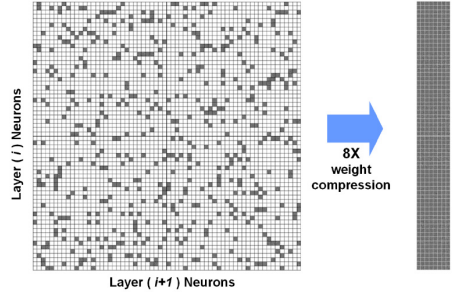

Fig. 2. Illustration of CGS. (Left) Weight matrix of 1024x1024 has 87.5% weights dropped with block-wise sparsity. (Right) Selected 16x16 weight blocks are compressed, stored with minimal index. (adapted from [17]).

## IV. PROPOSED SCHEME COMBINING LOW-PRECISION AND STRUCTURED SPARSITY

### A. Proposed Training Algorithm

Our proposed training algorithm is based on BNN with additional structured sparsity constraints of Coarse-Grain Sparsity (CGS). Prior to training, blocks of weights are randomly dropped off according to the CGS block size and CGS compression ratio. These blocks remain zero during training and inference. Training algorithm for non-sparse blocks of weights is similar to that of BNN training using back propagation.

#### 1) DNN with Coarse-Grained sparsity (DNN-CGS)

For fully-connected layers, the weight matrix is divided into square blocks. For $x*x$ CGS block size, each block contains $x^2$ weights. For convolution layers, $x*x$ CGS block size, contains $x^2$ 2D filters of size $l*w$, where $l$ is the length and $w$ is the width of the 2D filter. Once the weights are segregated into blocks, large number of blocks are randomly dropped off with probability equal to the CGS ratio. These blocks remain zero during training and inference and hence do not contribute to the physical memory. Fig. 2 shows an example weight matrix for a fully-connected layer of size 1024×1024, where each square represents a block of weights of size 16×16. Grey squares represent blocks where eligible connections are present and white squares represent blocks with absence of connections. Fig. 2 (right) illustrates the blocks with active connections, compressed along row, after applying 8X CGS ratio.

#### 2) Training of DNN-CGS with quantization

The sparse weight matrix/tensor, generated after applying CGS, is trained using back propagation. There are three phases of backpropagation algorithm: forward phase, backward phase and weight update phase.

During forward phase, quantized weights are generated from high-precision weights using 1-to-3-bit quantization function. Once the activations are computed, they are quantized. The quantized version of weights and activations is used for the forward pass. During backward phase, gradients of cost function with respect to activations and weights is computed starting from output to input layer. Straight-through estimator [12] is used to estimate gradient with respect to quantized activations. Fig. 3 (right) shows the estimated gradient for 2-bit quantization case. During weight update phase, the high precision weights are updated only for blocks of non-sparse weights, using Eq. (1).

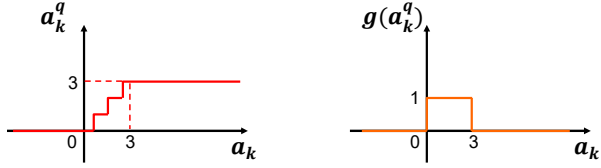

Fig. 3. Gradient estimation using straight-through estimator. (Left) 2-bit quatization example. (Right) estimated gradient.

$$(W_{ij})_{k+1} = (W_{ij})_k + \{(\Delta W_{ij})_k + m*(\Delta W_{ij})_{k-1}\}*lr*C_{ij} \quad (1)$$

where $(W_{ij})_k$ is weight in the weight matrix at $k^{th}$ iteration, $m$ is momentum, $lr$ is learning rate, and $C_{ij}$ is CGS connection coefficient between two consecutive DNN layers. $C_{ij} = 0$ for weights corresponding to the non-selected blocks and $C_{ij} = 1$ for weights corresponding to selected non-zero blocks.

### B. Proposed Hardware Design

The overall DNN acceleration system is shown in Fig. 4. The hardware supports two hidden layers with 512 neurons and 10 output layer neurons. Weights are stored in on-chip SRAM arrays. Each layer constitutes a set of MAC units followed by batch-norm and activation layer. In each hidden layer, input neurons are processed serially whereas the accumulation of weighted sum is done in parallel.

#### 1) Structured Sparsity and CGS decompression

Fully connected DNN memory is dominated by weights. To reduce the memory utilization, CGS-based compression is used to store the weights. To achieve structured compression, the neural network is trained by dividing DNN weights into blocks and randomly dropping them with a probability. Only the non-dropped weights with their corresponding index values are stored in on-chip SRAM arrays. Weight vectors are decompressed using demultiplexers by providing index bits as select signals.

#### 2) Batch-normalization

Conventional batch normalization follows Eq. (2) and Eq. (3) where we need to perform three additions, one multiplication and one division operation.

$$x' = \sum_i^n w_i \cdot a_i, \quad x = x' + b \quad (2)$$

$$y = \left(\frac{x-\mu}{\sigma}\right)\gamma + \beta \quad (3)$$

where $a_i$ is input activation, $w_i$ is weight, $b$ is bias, $x'$ is the weighted sum, $y$ is the output value before activation, and $\mu$ and $\sigma$ are the mean and standard deviation of the weighted sums in a batch, respectively, $\gamma$ and $\beta$ are batch-normalization scaling and shifting parameters.

These operations can be optimized by using new constants, namely, addition parameter $\beta'$ and multiplication parameter $\gamma'$. Then Eq. (2) can be reduced to Eq. (4), where only one multiplication and one addition operation is needed. This results in significant reduction in power and area.

$$y = x' \cdot \gamma' + \beta' \quad (4)$$

where $\beta' = \beta + \left(\frac{b-\mu}{\sigma}\right)\gamma$ and $\gamma' = \frac{\gamma}{\sigma}$.

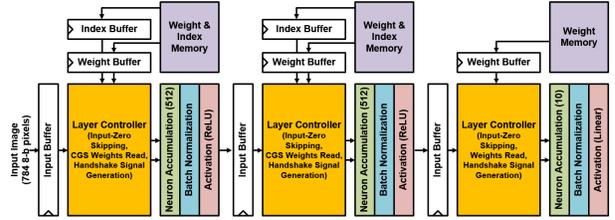

Fig. 4. Hardware architecture used for fully-connected DNNs.

#### 3) Zero skipping

By finding the active input neuron index, the zero skipping block skips the computation cycles for zero input activations. Active input neuron index is sent to on-chip compressed weight memory to fetch weights for all parallel MAC units. By exploiting the sparsity of input activations, latency can be reduced by 4.2X, on average, for the DNN using input images from the MNIST dataset.

#### 4) MAC unit for low precision weights

The accumulated values are passed to ReLU activation function to obtain output activations. For low precision weights (<3 bits), MAC multipliers are replaced by shifters, following the scheme in LightNN [16]. Each possible weight value is encoded with a signed shift value. For example, 2-bit weights are encoded as 00: -1/4, 01: -1/2, 10:1/2, 11:1/4 and multiplication of these weights with input activations was performed by shifters. High precision weights followed conventional MAC architecture.

#### 5) Pipeline control

The hardware consists of a feedforward architecture of hidden layers. All these layers were pipelined to increase the throughput. The number of cycles consumed by each layer depends on the number of non-zero input activations. Since the number of non-zero input values varies with input data, handshake signals are exchanged between adjacent layers to ensure proper execution of pipeline.

## V. EXPERIMENTAL RESULTS

### A. Experimental Setup

The proposed algorithm was used to train MNIST and CIFAR-10 datasets for image classification tasks. A Multi-Layer Perceptron (MLP) architecture with two hidden layers was used for training MNIST. Model selection for best architecture was performed based on accuracy study on different architectures. Test accuracy for different architecture settings have been compared in Fig. 5 for different models. Architectures with 1, 2 and 3 hidden layers with different number of neurons per hidden layer (128, 256, 512, 1024 neurons) are compared. Since the 2-hidden layer architecture with 512 neurons provides uncompromised test accuracy with less neurons/layers, this architecture is used for the MLP investigation for MNIST dataset.

For CIFAR-10 dataset, we employed the CNN architecture inspired by VGG [1], which was used in BNN [5]. This CNN consists of 6-covolution, 3 max-pooling and 3-fully connected layers (C128-C128-P2-C256-C256-P2-C512-C512-P2-F1024-F1024-F10). The architecture uses 3×3 convolution filters and batch-normalization layers. Deep learning framework Theano and toolbox Lasagne is used for training and testing of the models.

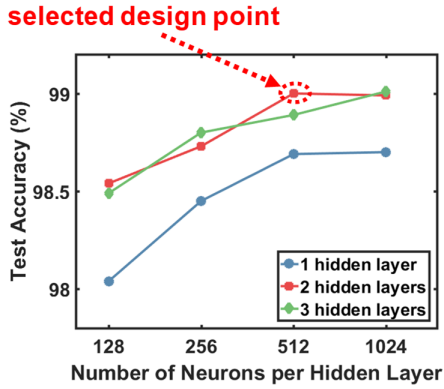

Fig. 5. Design point selection of MLP for MNIST dataset.

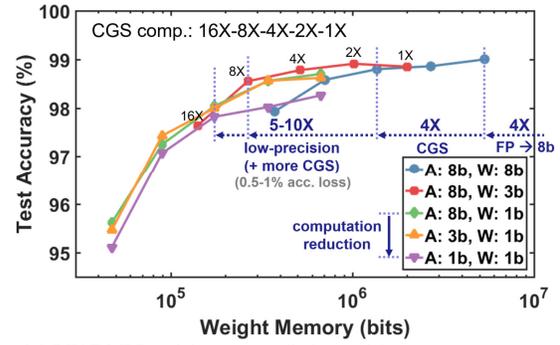

Fig. 6. MNIST MLP weight memory (in log scale) versus test accuracy.

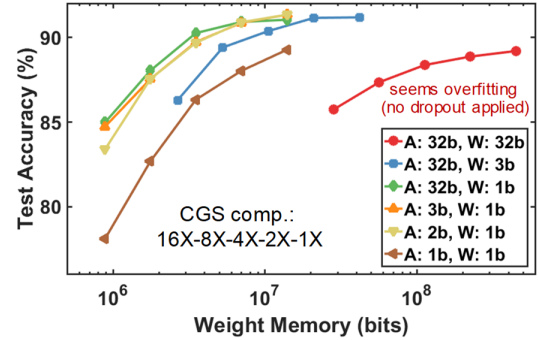

Fig. 7. CIFAR-10 CNN weight memory (in log scale) versus test accuracy.

For MNIST MLP, we investigated 5 different activation (A) and weight (W) precision settings: (A:8b, W:8b), (A:8b, W:3b), (A:8b, W:1b), (A:3b, W:1b), (A:1b, W:1b), from floating point (8-bit) down to 1-bit. For CIFAR-10 CNN, we investigated 6 different activation (A) and weight (W) precision settings: (A:32b, W:32b), (A:32b, W:3b), (A:32b, W:1b), (A:3b, W:1b), (A:2b, W:1b), (A:1b, W:1b), from floating point (32-bit) down to 1-bit. CGS block size for fully-connected layers is 16×16 and on convolution layers is 4×4. Test is performed for 5 different CGS compression ratios: 16X-8X-4X-2X-1X, where percentage of non-sparse weight blocks are 6.25%-12.5%-25%-50%-100%, respectively. CGS is not applied to the output layer as it has small number of weights. The activation quantization are essentially quantized values of the ReLU activation. For example, 1-bit activations are 0 or +1; 2-bit activations are 0, +1, +2, +3 and 3-bit activations are 0, +0.25, +0.5, +0.75, +1, +1.25, +1.5, or +1.75. The quantization levels are equally spaced. Activation quantization for the 2-bit case is shown in Fig. 3 (left). The quantized valued of weights is discussed in Section IV.B.

The hardware architecture was implemented in TSMC 40nm LP CMOS with high $V_t$ devices. All the designs, with different weight and activation precisions, are synthesized at 100MHz with extensive clock gating. DNN weights are stored in SRAM arrays generated from a commercial memory compiler. For each design with different weight precisions, new SRAM arrays were generated such that 512 weights fit in one row. Test accuracy and latency are obtained from post-synthesis simulation for the entire MNIST test dataset of 10k images. Power numbers are obtained from Synopsys Primetime PX using data activity of fully connected DNN layers' post-synthesis netlist.

### B. Algorithm Results and Analysis

Comparison of test accuracy versus weight memory on MNIST for different models is shown in Fig. 6. The weight memory includes the index of CGS blocks. Activation precision does not affect memory requirement as activations are not stored. As indicated in the plot certain regions show high level of sparsity with very low accuracy degradation. In Fig. 6, 4X compression on high precision model results in accuracy loss of only 0.20%. DNN model with 8X compression and 3-bit weight quantization (10X weight memory reduction) shows minimal accuracy degradation of 0.45% compared to high precision and uncompressed network. Even highly quantized network (A:3b, W:1b) and 2X compressed DNN shows minimal accuracy degradation (~0.40%). Computation saving on this model will be more pronounced as quantization of activations is not reflected in the weight memory saving. With maximally quantized networks (A:1b, W:1b) and 2X compression, accuracy degradation of ~1% might be tolerable considering that this reduces the overall memory by ~16X. With CGS compression rate more than 10X on 1-bit weight models, accuracy degradation becomes conspicuous (2-4%).

Fig. 7 shows the accuracy versus weight memory trade-off for the CNN for CIFAR-10 dataset. It can be seen that the floating-point DNN actually has lower accuracy than that of compressed networks, seemingly due to overfitting. This observation is in line with the argument made in [4] that quantization noise acts like a regularizer. Even highly quantized (A:2b, W:1b) and compressed (CGS: 4X) network shows 0.49% higher accuracy compared to floating-point network. Similar to the observation for MNIST, for highly quantized network (A:1b, W:1b), accuracy degradation is pronounced when CGS compression is applied. Judicious selection of DNN quantization (most suitable setting is A:3b, W:1b) and structured CGS compression of up to 4X can give similar and at times better accuracy, compared to floating point and uncompressed network.

### C. Hardware Results and Analysis

Fig. 8 shows the accuracy and post-synthesis area of memory and logic for different weight/activation precisions and CGS ratios. When using higher CGS compression ratio and lower precision, the memory area significantly decreases and the logic area starts dominating the total area. Reducing the weight precision is more effective for the overall area reduction, compared to lowering the activation precision. The smallest area of 0.47mm$^2$ is achieved by the BNN design (1-bit activation and 1-bit weight) with 8X CGS compression.

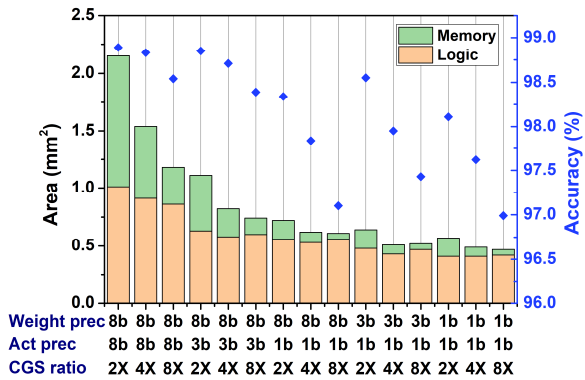

Fig. 8. Area breakdown for different combination of weight precision, activation precision and CGS compression ratio.

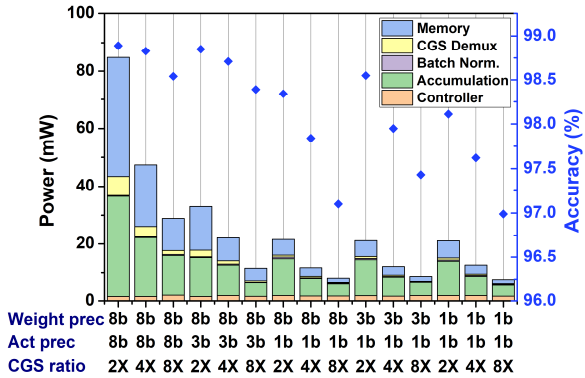

Fig. 9. Power breakdown for different combination of weight percision, activation precision and CGS compression ratio.

Fig. 9 shows the accuracy and post-synthesis power breakdown of the competing designs. Compressing the memory using CGS significantly reduces the power consumption by reducing the memory size and number of weight accumulations. The total logic power is dominated by accumulation of weighted sum. This exceeds the memory power which has been reduced due to aggressive weight compression. Fig. 10 shows the energy per image and the classification accuracy tradeoff for different activation/weight precision values and CGS compression ratios.

With 8-bit activations, 3-bit weights, and 8X compression, 98.4% MNIST accuracy was achieved with 20nJ energy per classification, which is a favorable accuracy-energy trade-off compared to much lower precision DNNs with less compression (BNN achieves 13nJ energy at 97% accuracy).

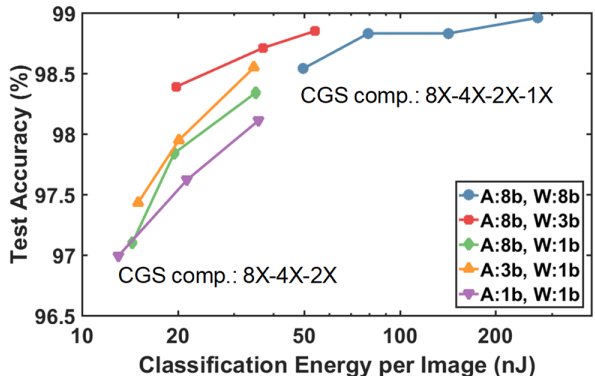

Fig. 10. Classification energy and test accuracy of MNIST MLP designs with different precision and structured compression.

Compared to the uncompressed DNN with 8-bit precision, we achieve >10X energy reduction with only 0.6% accuracy loss, by optimal combining of low precision and CGS compression.

## VI. CONCLUSION

In this work, we have studied joint-optimization of low-precision and structured sparsity towards DNN designs with minimal area and energy, by co-designing algorithm and hardware. We have shown 10-50X weight memory reduction on MLP for MNIST dataset and CNN for CIFAR-10 dataset, compared to floating-point DNN counterparts, with minimal accuracy degradation (<0.5%). We have presented analysis on optimized combination of very low precision and structured compression for favorable energy, area, and accuracy tradeoffs, based on a number of DNN implementations in 40nm LP CMOS. The MLP DNN designed with 8-bit activations, 3-bit weights, and 8X structured compression showed 98.4% accuracy at 20nJ energy per classification, outperforming further lower precision designs with less structured compression.

ACKNOWLEDGEMENT

This work was supported in part by Intel Labs, NSF grants 1652866, 1715443, 1740225 and Office of Naval Research grants N000141410722, N000141712826.